\documentclass[runningheads]{llncs}
\usepackage[T1]{fontenc}

\usepackage{multirow}
\usepackage{epsfig}
\usepackage{amsmath}
\usepackage{amsfonts}
\usepackage{float}
\usepackage{booktabs}
\usepackage{color}
\usepackage[colorlinks]{hyperref}

\usepackage{subfigure}

\usepackage{marvosym}
\usepackage{ifsym}
\usepackage{graphicx}
\begin{document}
\title{UOD: Universal One-shot Detection of Anatomical Landmarks}


%
%

\author{
Heqin Zhu\inst{1,2,3} \and Quan Quan\inst{3} \and Qingsong Yao\inst{3} \and Zaiyi Liu\inst{4,5} \and S. Kevin Zhou\inst{1,2}$^{\href{mailto:s.kevin.zhou@gmail.com}{\textrm{\Letter}}}$
}

\authorrunning{H. Zhu et al.}

\institute{\
School of Biomedical Engineering, Division of Life Sciences and Medicine, University of Science and Technology of China, Hefei, Anhui, 230026, P.R.China \\
\and Suzhou Institute for Advanced Research, University of Science and Technology of China, Suzhou, Jiangsu, 215123, P.R.China
\\
\and Key Lab of Intelligent Information Processing of Chinese Academy of Sciences (CAS), Institute of Computing Technology, CAS, Beijing 100190, China\\
\and Department of Radiology, Guangdong Provincial People's Hospital, Guangdong Academy of Medical Sciences, Guangzhou, China \\
\and Guangdong Provincial Key Laboratory of Artificial Intelligence in Medical Image Analysis and Application, Guangdong Provincial People's Hospital, Guangdong Academy of Medical Sciences, Guangzhou, China 
}

\maketitle           

\begin{abstract}
One-shot medical landmark detection gains much attention and achieves great success for its label-efficient training process. However, existing one-shot learning methods are highly specialized in a single domain and suffer domain preference heavily in the situation of multi-domain unlabeled data. Moreover, one-shot learning is not robust that it faces performance drop when annotating a sub-optimal image. To tackle these issues, we resort to developing a domain-adaptive one-shot landmark detection framework for handling multi-domain medical images, named \textbf{Universal One-shot Detection (UOD)}. UOD consists of two stages and two corresponding universal models which are designed as combinations of domain-specific modules and domain-shared modules. In the first stage, a domain-adaptive convolution model is self-supervised learned to generate pseudo landmark labels. In the second stage, we design a domain-adaptive transformer to eliminate domain preference and build the global context for multi-domain data. Even though only one annotated sample from each domain is available for training, the domain-shared modules help UOD aggregate all one-shot samples to detect more robust and accurate landmarks. We investigated both qualitatively and quantitatively the proposed UOD on three widely-used public X-ray datasets in different anatomical domains (i.e., head, hand, chest) and obtained state-of-the-art performances in each domain. The code is at \href{https://github.com/heqin-zhu/UOD_universal_oneshot_detection}{https://github.com/heqin-zhu/UOD\_universal\_oneshot\_detection}.

\end{abstract}

\section{Introduction}
Robust and accurate detecting of anatomical landmarks is an essential task in medical image applications~\cite{zhou2019handbook,zhou2021review}, which plays vital parts in varieties of clinical treatments, for instance, vertebrae localization~\cite{wang2022accurate}, orthognathic and orthodontic surgeries~\cite{lang2022dentalpointnet}, and craniofacial anomalies assessment~\cite{elkhill2022graph}. Moreover, anatomical landmarks exert their effectiveness in other medical image tasks such as segmentation~\cite{chen2021automated}, registration~\cite{espinel2021using}, and biometry estimation~\cite{avisdris2022biometrynet}.

In the past years, lots of fully supervised methods~\cite{zhu2022learning,zhu2021you,jiang2022cephalformer,payer2019integrating,wang2022accurate,elkhill2022graph,yao2020miss,payer2019integrating} have been proposed to detect landmarks accurately and automatically. To relieve the burden of experts and reduce the amount of annotated labels, various one-shot and few-shot methods have been come up with. Zhao et al.~\cite{zhao2019data} demonstrate a model which learns transformations from the images and uses the labeled example to synthesize additional labeled examples, where each transformation is composed of a spatial deformation field and an intensity change. Yao et al.~\cite{yao2021one} develop a cascaded self-supervised learning framework for one-shot medical landmark detection. They first train a matching network to calculate the cosine similarity between features from an image and a template patch, then fine-tune the pseudo landmark labels from coarse to fine. Browatzki et al.~\cite{browatzki20203fabrec} propose a semi-supervised method that consists of two stages. They first employ an adversarial auto-encoder to learn implicit face knowledge from unlabeled images and then fine-tune the decoder to detect landmarks with few-shot labels.

However, one-shot methods are not robust enough because they are dependent on the choice of labeled template and the accuracy of detected landmarks may decrease a lot when choosing a sub-optimal image to annotate. To address this issue, Quan et al.~\cite{quan2022images} propose a novel Sample Choosing Policy (SCP) to select the most worthy image to annotate. Despite the improved performance, SCP brings an extra computation burden. Another challenge is the scalability of model building when facing multiple domains (such as different anatomical regions). While conventional wisdom is to independently train a model for each domain, Zhu et al.~\cite{zhu2021you} propose a universal model YOLO for detecting landmarks across different anatomies and achieving better performances than a collection of single models. YOLO is regularly supervised using the CNN as backbone and it is unknown if the YOLO model works for a one-shot scenario and with a modern transformer architecture. 

Motivated by above challenges, to detecte robust multi-domain label-efficient landmarks, we design domain-adaptive models and propose a universal one-shot landmark detection framework called \textbf{Universal One-shot Detection (UOD)}, illustrated in Figure~\ref{fig_overall}. A universal model is comprised of domain-specific modules and domain-shared modules, learning the specified features of each domain and common features of all domains to eliminate domain preference and extract representative features for multi-domain data. Moreover, one-shot learning is not robust enough because of the sample selection while multi-domain one-shot learning reaps benefit from different one-shot samples from various domains, in which cross-domain features are excavated by domain-shared modules. Our proposed UOD framework consists of two stages: 1) Contrastive learning for training a universal model with multi-domain data to generate pseudo landmark labels. 2) Supervised learning for training domain-adaptive transformer (DATR) to avoid domain preference and detect robust and accurate landmarks.

In summary, our contributions can be categorized into three parts: \textbf{1)} We design the first universal framework for multi-domain one-shot landmark detection, which improves detecting accuracy and relieves domain preference on multi-domain data from various anatomical regions. \textbf{2)} We design a domain-adaptive transformer block (DATB), which is effective for multi-domain learning and can be used in any other transformer network. \textbf{3)} We carry out comprehensive experiments to demonstrate the effectiveness of UOD for obtaining SOTA performance on three publicly used X-ray datasets of head, hand, and chest.

\section{Method}

\begin{figure}[t]
    \centering
    \includegraphics[width=0.95\textwidth]{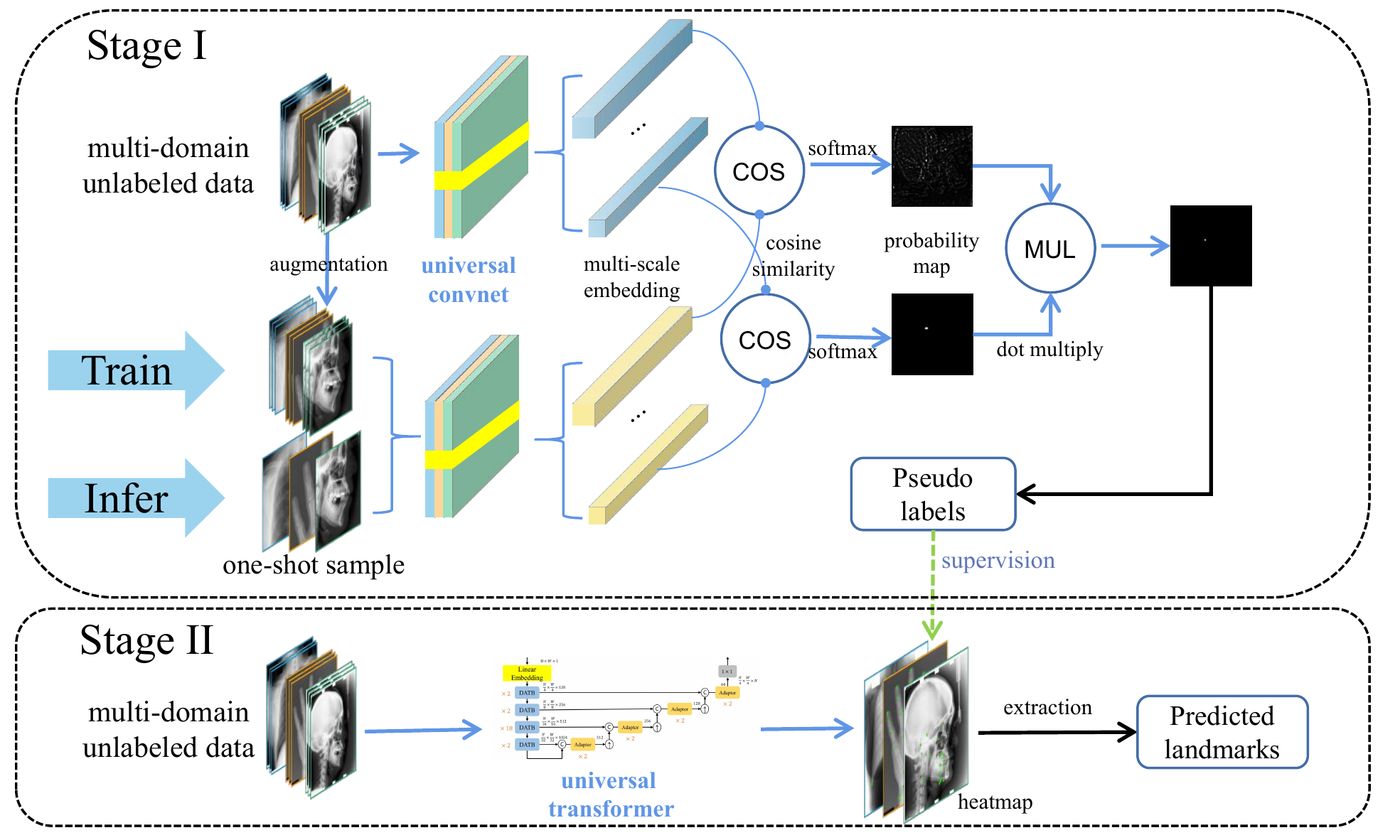}
    \caption{Overview of UOD framework. In stage I, two universal models are learned via contrastive learning for matching similar patches from original image and augmented one-shot sample image and generating pseudo labels. In stage II, DATR is designed to better capture global context information among all domains for detecting more accurate landmarks.}
    \label{fig_overall}
\end{figure}

As Figure~\ref{fig_overall} shows, UOD consists of two stages: 1) Contrastive learning and 2) Supervised learning. In stage I, to learn the local appearance of each domain, a universal model is trained via self-supervised learning, which contains domain-specific VGG~\cite{vgg} and UNet~\cite{ref_unet} decoder with each standard convolution replaced by a domain adaptor~\cite{ref_u2net}. In stage II, to grasp the global constraint and eliminate domain preference, we designed a domain-adaptive transformer (DATR). 

\subsection{Stage I: Contrastive learning}
As Figure~\ref{fig_overall} shows, following Yao et al.~\cite{yao2021one}, we employ contrastive learning to train siamese network for matching similar patches of original image and augmented image. Given a multi-domain input image $X^d\in R^{H^d\times W^d \times C^d}$ belongs to domain $d$ from multi-domain data, we randomly select a target point $P$ and crop a half-size patch $X_p^d$ which contains $P$. After applying data augmentation on $X_p^d$, the target point is mapped to $P_p$. Then we feed $X^d$ and $X_p^d$ into the siamese network respectively and obtain the multi-scale feature embeddings. We compute cosine similarity of two feature embeddings from each scale and apply $\text{softmax}$ to the cosine similarity map to generate a probability matrix. Finally, we calculate the cross entropy loss of the probability matrix and ground truth map which is produced with the one-hot encoding of $P_p^d$ to optimize the siamese network for learning the latent similarities of patches. At inferring stage, we replace augmented patch $X_p^d$ with the augmented one-shot sample patch $X_s^d$. We use the annotated one-shot landmarks as target points to formulate the ground truth maps. After obtaining probability matrices, we apply $\mathop{\arg\max}$ to extract the strongest response points as the pseudo landmarks, which will be used in UOD Stage II. 

\subsection{Stage II: Supervised learning}
In stage II, we design a universal transformer to capture global relationship of multi-domain data and train it with the pseudo landmarks generated in stage I. The universal transformer has a domain-adaptive transformer encoder and domain-adaptive convolution decoder. The decoder is based on a U-Net~\cite{ref_unet} decoder with each standard convolution replaced by a domain adaptor~\cite{ref_u2net}. The encoder is based on Swin Transformer~\cite{swin} with shifted window and limited self-attention within non-overlapping local windows for computation efficiency. Different from Swin Transformer~\cite{swin}, we design a domain-adaptive transformer block (DATB) and use it to replace the original transformer block.

\subsubsection{Domain-adaptive transformer encoder}\label{sec:encoder}

\begin{figure}[t]
\centering
    \includegraphics[width=0.95\textwidth]{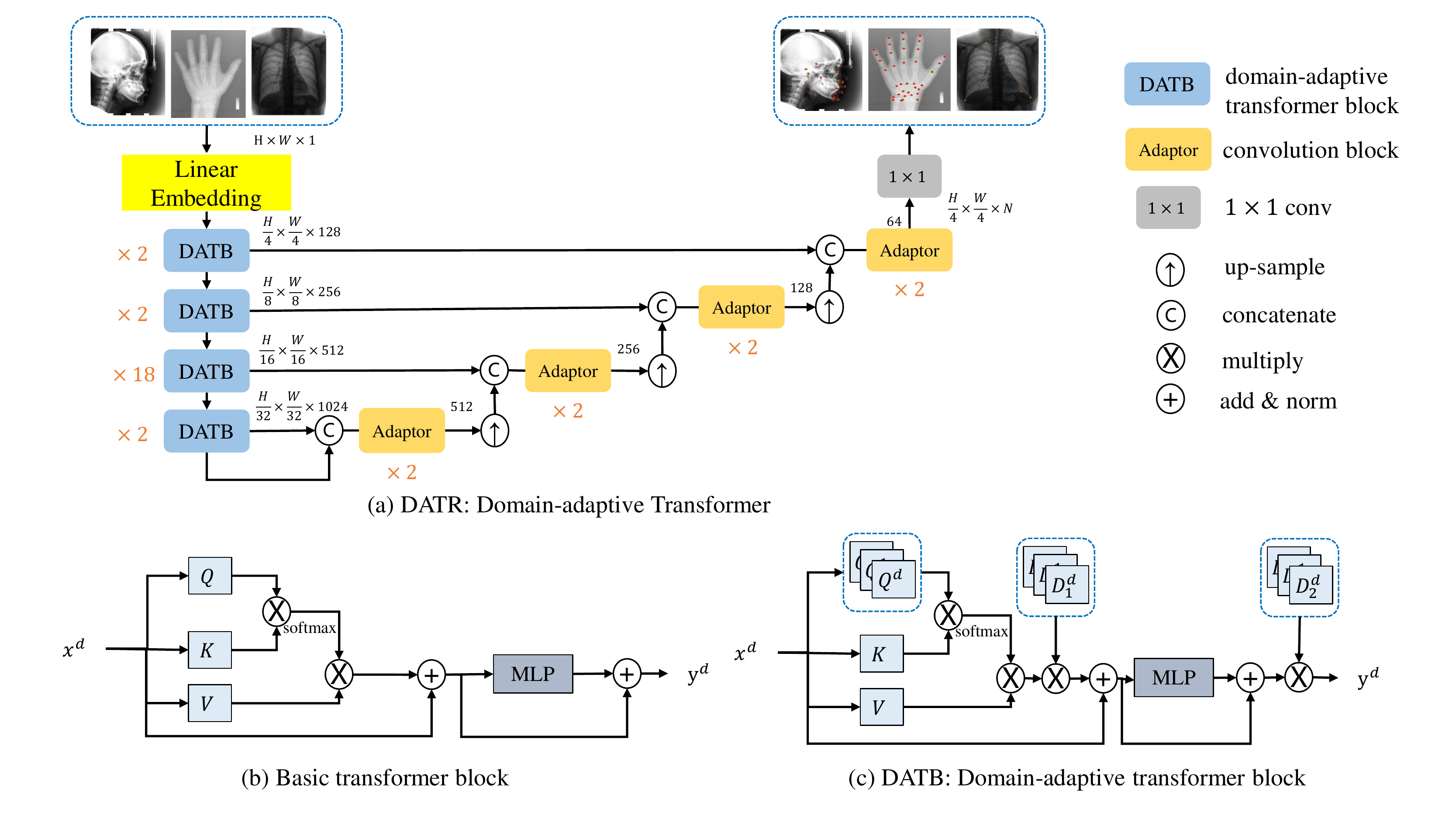}
    \caption{(a) The architecture of DATR in stage II, which is composed of domain-adaptive transformer encoder and convolution adaptors~\cite{ref_u2net}. (b) Basic transformer block. (c) Domain-adaptive transformer block. Each domain-adaptive transformer is a basic transformer block with query matrix duplicated and domain-adaptive diagonal for each domain. The batch-normalization, activation, and patch merging are omitted.}
    \label{fig_datr}
\end{figure}

As Figure~\ref{fig_datr}(a) shows, the transformer encoder is built up with DATB, making full use of the capability of transformer for modeling global relationship and extracting multi-domain representative features. As in Figure~\ref{fig_datr}(b), a basic transformer block~\cite{Transformer} consists of a multi-head self-attention module (MSA), followed by a two-layer MLP with GELU activation. Furthermore, layer normalization (LN) is adopted before each MSA and MLP and a residual connection is adopted after each MSA and MLP. Given a feature map $x^d \in R^{h\times w\times c}$ from domain $d$ with height $h$, width $w$, and $c$ channels, the output feature maps of MSA and MLP, denoted by $\hat{y}^d$ and $y^d$, respectively, are formulated as:
\begin{equation}
    \begin{split}
        \hat{y}^d &=  \text{MSA}(\text{LN}(x^d))+x^d\\
        y^d &= \text{MLP}(\text{LN}(\hat{y}^d)) + \hat{y}^d 
    \end{split}
    \label{Eq:trans}
\end{equation}
where $\text{MSA} = \text{softmax}(QK^T)V$.

As illustrated in Figure~\ref{fig_datr}(b)(c), DATB is based on Eq.~(\ref{Eq:trans}). Similar to U2Net~\cite{ref_u2net} and GU2Net~\cite{zhu2021you}, we adopt domain-specific and domain-shared parameters in DATB. Since the attention probability is dependent on query and key matrix which are symmetrical, we duplicate the query matrix for each domain to learn domain-specific query features and keep key and value matrix domain-shared to learn common knowledge and reduce parameters. Inspired by LayerScale~\cite{layerscale}, we further adopt learnable diagonal matrix~\cite{layerscale} after each MSA and MLP module to facilitate the learning of domain-specific features, which costs few parameters ($O(N)$ for $N\times N$ diagonal). Different from LayerScale~\cite{layerscale}, proposed domain-adaptive diagonal $D_1^d$ and $D_2^d$ are applied for each domain with $D_2^d$ applied after residual connection for generating more representative and direct domain-specific features. The above process can be formulated as:
\begin{equation}
    \begin{split}
        \hat{y}^d &=  D_1^d \times \text{MSA}_{Q^d}(\text{LN}(x^d))+x^d\\
        y^d &= D_2^d \times(\text{MLP}(\text{LN}(\hat{y}^d)) + \hat{y}^d )
    \end{split}
\end{equation}
where $\text{MSA}_{Q^d} = \text{softmax}(Q^dK^T)V$.

\subsubsection{Overall pipeline} \label{sec:overall}

Given that a random input $X^d\in R^{H^d\times W^d \times C^d}$ belongs to domain $d$ from mixed datasets on various anatomical regions, which contains $N^d$ landmarks with corresponding coordinates being $\{(i_1^d,j_1^d), (i_2^d,j_2^d), \ldots, (i_{N_d}^d,j_{N_d}^d)\}$, we set the $n\text{-th}\in\{1,2,\dots,N^d\}$ initial heatmap $\tilde{Y}_{n}^d \in R^{H^d\times W^d \times C^d }$ with Gaussian function to be $\tilde{Y}_{n}^d= \frac{1}{\sqrt{2\pi}\sigma}e^{-\frac{(i-i_{n}^d)^2+(j-j_{n}^d)^2}{2\sigma^2}}$ if $\sqrt{(i-i_{n}^d)^2+(j-j_{n}^d)^2} \le \sigma$ and 0 otherwise. We further add an exponential weight to the Gaussian distribution to distinguish close heatmap pixels and obtain the ground truth heatmap $Y_{n}^d(i,j) = \alpha ^{\tilde{Y}_{n}^d(i,j)}$.

As illustrated in Figure~\ref{fig_datr}, firstly, the input image from a random batch is partitioned into non-overlapping patches and linearly embedded. Next, these patches are fed into cascaded transformer blocks at each stage, which are merged except in the last stage. Finally, a domain-adaptive convolution decoder makes dense prediction to generate heatmaps, which is further used to extract landmarks via threshold processing and connected components filtering. 

\section{Experiment}

\textbf{Datasets.} For performance evaluation, we adopt three public X-ray datasets from different domains on various anatomical regions of head, hand, and chest. (i) \underline{Head dataset} is a widely-used dataset for IEEE ISBI 2015 challenge~\cite{ref_head,wang2015evaluation} which contains 400 X-ray cephalometric images with 150 images for training and 250 images for testing. Each image is of size $2400\times 1935$ with a resolution of $0.1mm \times 0.1mm$, which contains 19 landmarks manually labeled by two medical experts and we use the average labels same as Payer et al.~\cite{payer2019integrating}. (ii) \underline{Hand dataset} is collected by~\cite{gertych2007bone} which contains 909 X-ray images and 37 landmarks annotated by~\cite{payer2019integrating}. We follow ~\cite{zhu2021you} to split this dataset into a training set of 609 images and a test set of 300 images. Following \cite{payer2019integrating} we assume the distance between two endpoints of wrist is 50mm and calculate the physical distance as $\text{distance}_{\text{physical}}=\text{distance}_{\text{pixel}} \times \frac{50}{\|p-q\|_2}$  where $p,q$ are the two endpoints of the wrist respectively. (iii) \underline{Chest dataset}~\cite{zhu2021you} is a popular chest radiography database collected by Japanese Society of Radiological Technology (JSRT)~\cite{shiraishi2000development} which contains 247 images. Each image is of size $2048\times 2048$ with a resolution of $0.175mm \times 0.175mm$. We split it into a training set of 197 images and a test set of 50 images and select 6 landmarks from landmark labels at the boundary of the lung as target landmarks. 

\noindent \textbf{Implementation details.}
UOD is implemented in Pytorch and trained on a TITAN RTX GPU with CUDA version being 11. All encoders are initialized with corresponding pre-trained weights. We set batch size to 8, $\sigma$ to 3, and $\alpha$ to 10. We adopt binary cross-entropy (BCE) as loss function for both stages. In stage I, we resize each image to the same shape of $384 \times 384$ and train universal convolution model by Adam optimizer for 1000 epochs with a learning rate of 0.00001. In stage II, we resize each image to the same shape of $576\times 576$ and optimize the universal transformer by Adam optimizer for 300 epochs with a learning rate of 0.0001. When calculating metrics, all predicted landmarks are resized back to the original size. For evaluation, we choose model with minimum validation loss as the inference model and adopt two metrics: mean radial error (MRE) $\text{MRE} = \frac{1}{N}\sum_{i}^{N}\sqrt{(x_i-\tilde{x}_i)^2 +(y_i-\tilde{y}_i)^2}$ and successful detection rates (SDR) within different thresholds $t$: $\text{SDR}(t) = \frac{1}{N}\sum_{i}^{N}\delta(\sqrt{(x_i-\tilde{x}_i)^2 +(y_i-\tilde{y}_i)^2}\le t)$.

\subsection{Experimental results}
\begin{figure}[t]
	\centering
	\begin{minipage}{0.45\linewidth}
		\centering
        \includegraphics[width=\linewidth]{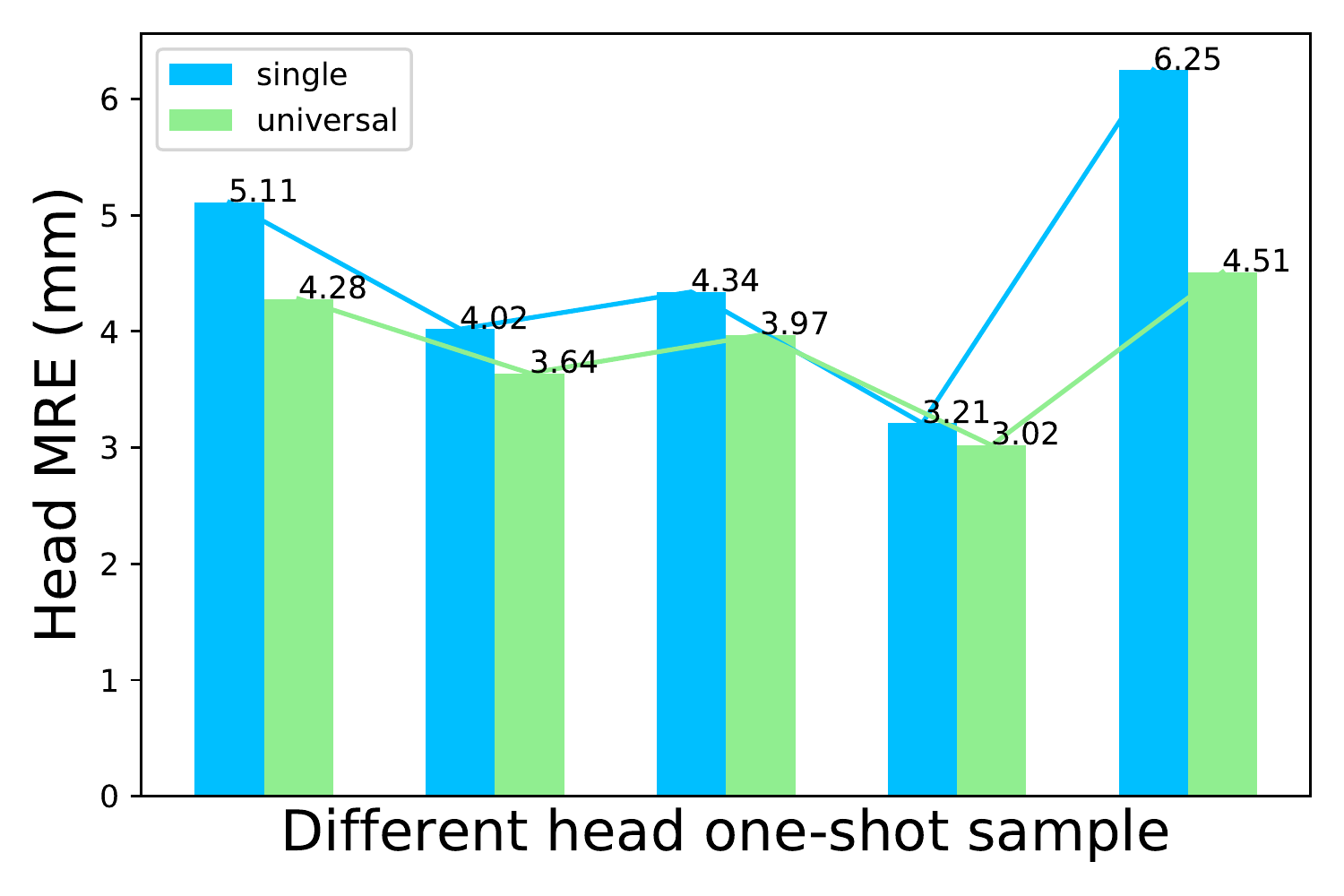}
		\label{fig:id_head_mre}
	\end{minipage}
	\begin{minipage}{0.45\linewidth}
        \centering
        \includegraphics[width=\linewidth]{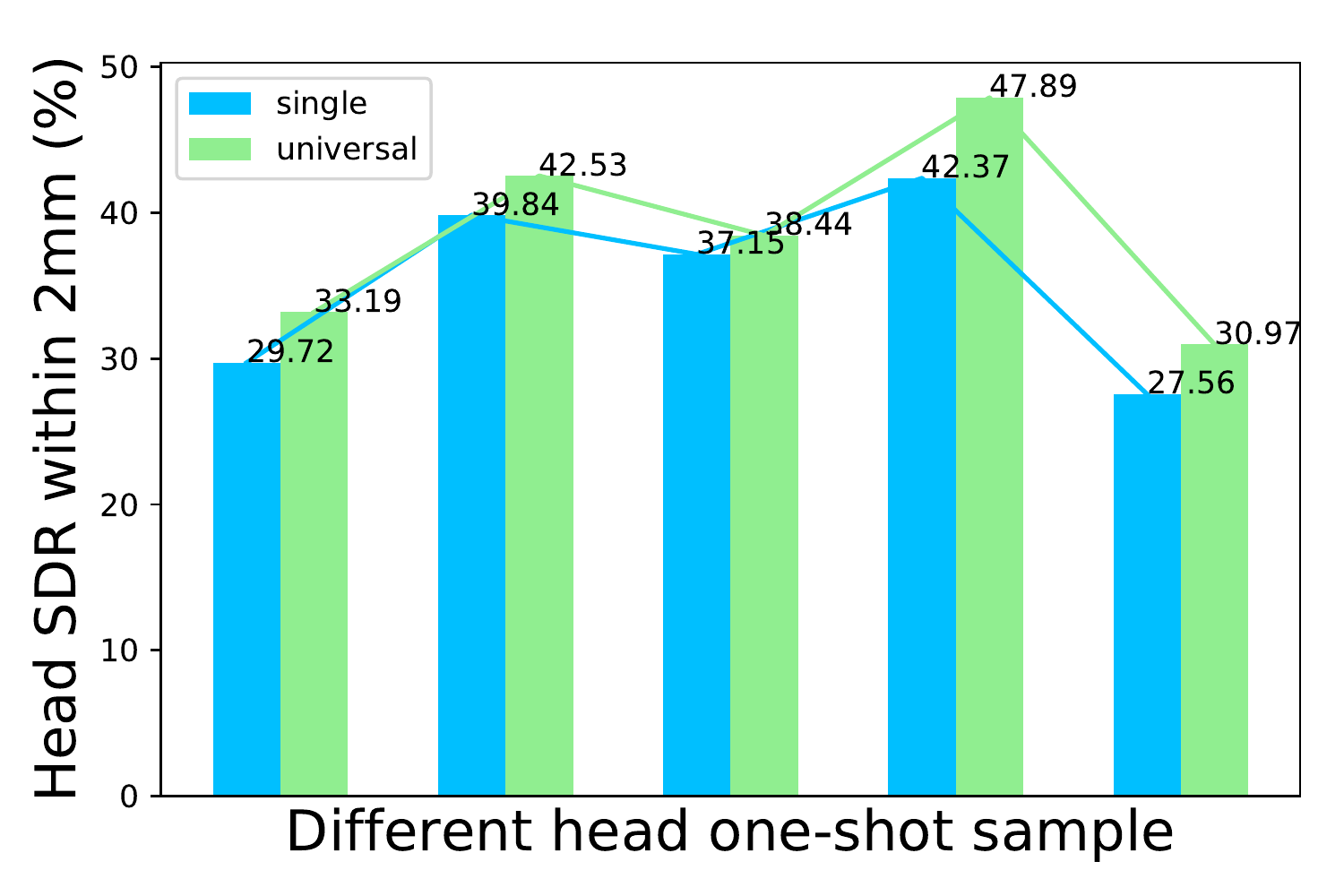}
		\label{fig:id_head_sdr}
	\end{minipage}
\caption{Comparison of single model and universal model on head dataset.}
\label{fig:id_head}
\end{figure}

\noindent{\textbf{The effectiveness of universal model:}}
To demonstrate the effectiveness of universal model for multi-domain one-shot learning, we adopt head and hand datasets for evaluation. In stage I, the convolution models are trained in two ways: 1) single: trained on every single dataset respectively, and 2) universal: trained on mixed datasets together. With a fixed one-shot sample for the hand dataset, we change the one-shot sample for the head dataset and report the MRE and SDR of the head dataset. As Figure~\ref{fig:id_head} shows, universal model performs much better than single model on various one-shot samples and metrics. It is proved that universal model learns domain-shared knowledge and promotes domain-specific learning. Furthermore, the MRE and SDR metrics of universal model have a smaller gap among various one-shot samples, which demonstrates the robustness of universal model learned on multi-domain data.

\begin{table}[t]
\centering
\caption{Quantitative comparison of UOD with SOTA methods on head, hand, and chest datasets. * denotes the method is trained on every single dataset respectively while \dag denotes the method is trained on mixed data.}
\label{tab_results}
\resizebox{1\linewidth}{!}{
\begin{tabular}{lrrrrrrrrrrrrrr}
\hline
\multirow{3}*{Method} & &\multicolumn{5}{c}{Head~\cite{ref_head}}  &\multicolumn{4}{c}{Hand~\cite{gertych2007bone}}  &\multicolumn{4}{c}{Chest~\cite{shiraishi2000development}}  \\
&Label  &MRE$\downarrow$ &\multicolumn{4}{c}{SDR$\uparrow$ (\%)}  &MRE$\downarrow$ &\multicolumn{3}{c}{SDR$\uparrow$ (\%)}  &MRE$\downarrow$ &\multicolumn{3}{c}{SDR$\uparrow$ (\%)}\\
& &(mm)& 2mm   & 2.5mm & 3mm   & 4mm   &(mm)     & 2mm & 4mm & 10mm  &(mm)  & 2mm & 4mm & 10mm\\
\hline
YOLO~\cite{zhu2021you}\dag &all      &1.32 &81.14 &87.85 &92.12 &96.80  &0.85 &94.93 &99.14&99.67  &4.65 &31.00 &69.00 &93.67 \\
\hline
YOLO~\cite{zhu2021you}\dag &25      &1.96 &62.05 &77.68 &88.21 &97.11  &2.88 &72.71 &92.32&97.65  &7.03 &19.33 &51.67 &89.33 \\
\hline
YOLO~\cite{zhu2021you}\dag &10      &2.69 &47.58 &66.47 &78.42 &90.89  &9.70 &48.66 &76.69&90.52  &16.07 &11.67 &33.67 &76.33    \\
\hline
YOLO~\cite{zhu2021you}\dag &5      &5.40 &26.16 &41.32 &54.42 &73.74  &24.35 &20.59 &48.91 & 72.94  &34.81 &4.33 &19.00 &56.67 \\
\hline
\hline

CC2D~\cite{yao2021one}*  &1     &2.76 &42.36 &51.82 &64.02 &78.96  &2.65 &51.19 &82.56 &95.62 &10.25 &11.37&35.73&68.14\\
\hline
Ours\dag&1  &\textbf{2.43} &\textbf{51.14}&\textbf{62.37}&\textbf{74.40}&\textbf{86.49}  &\textbf{2.52} &\textbf{53.37}&\textbf{84.27}&\textbf{97.59} &\textbf{8.49} &\textbf{14.00}&\textbf{39.33}&\textbf{76.33}\\

\hline
\end{tabular}
}
\end{table}

\begin{figure*}[t]
    \centering 
   
    \begin{minipage}[t]{0.16\textwidth}
        \centering
        \includegraphics[width=1\textwidth]{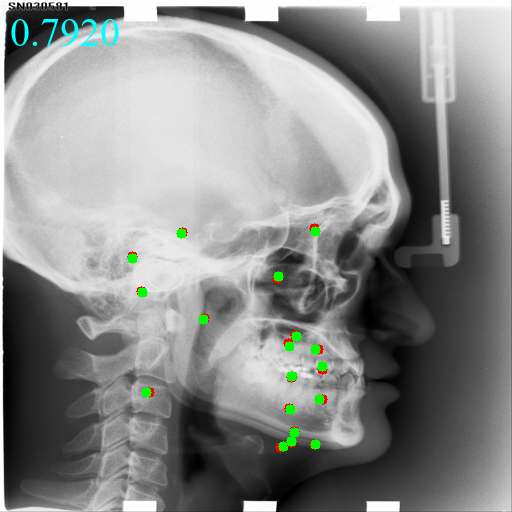}
        UOD
    \end{minipage}
        \begin{minipage}[t]{0.16\textwidth}
        \centering
        \includegraphics[width=1\textwidth]{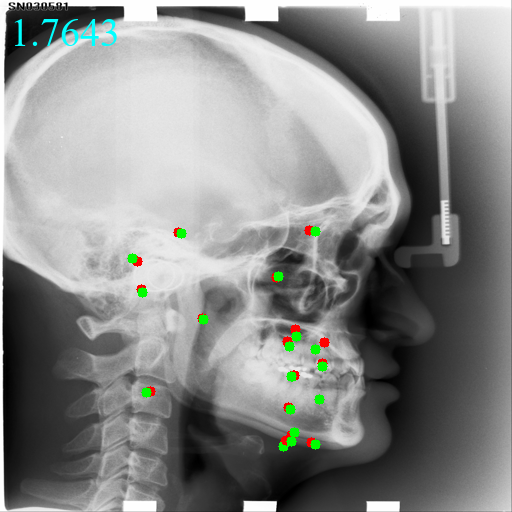}
        CC2D
    \end{minipage}
    \begin{minipage}[t]{0.16\textwidth}
        \centering
        \includegraphics[width=1\textwidth]{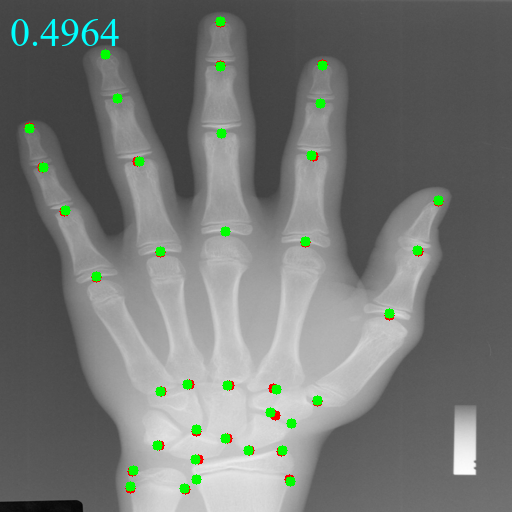}
        UOD
    \end{minipage}
        \begin{minipage}[t]{0.16\textwidth}
        \centering
        \includegraphics[width=1\textwidth]{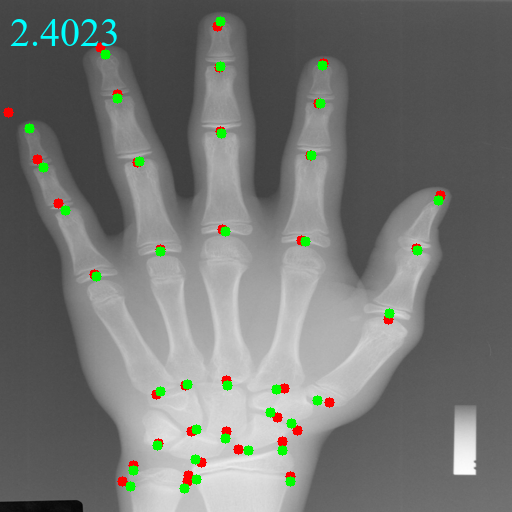}
        CC2D
    \end{minipage}
    \begin{minipage}[t]{0.16\textwidth}
        \centering
        \includegraphics[width=1\textwidth]{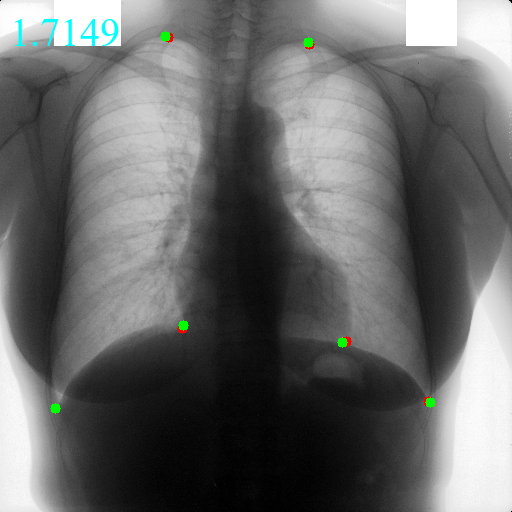}
        UOD
    \end{minipage}
    \begin{minipage}[t]{0.16\textwidth}
        \centering
        \includegraphics[width=1\textwidth]{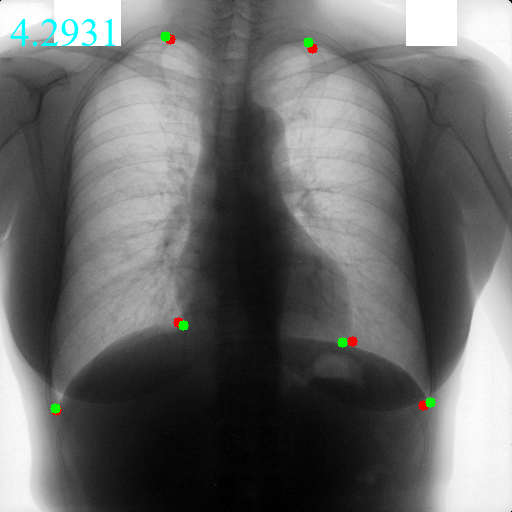}
        CC2D
    \end{minipage}
      
    \caption{Qualitative comparison of UOD and CC2D~\cite{yao2021one} on head, hand, and chest datasets. The red points \textcolor{red}{$\bullet$} indicate predicted landmarks while the green points \textcolor{green}{$\bullet$} indicate ground truth landmarks. The MRE value is displayed in the top left corner of the image.}
    \label{fig_visual}
\end{figure*}

\noindent{\textbf{Comparisons with state-of-the-art methods:}}
As Table~\ref{tab_results} shows, we compare UOD with two open-source landmark detection methods, i.e., YOLO~\cite{zhu2021you} and CC2D~\cite{yao2021one}. YOLO is a multi-domain supervised method while CC2D is a single-domain one-shot method. UOD achieves SOTA results on all datasets under all metrics, outperforming the other one-shot method by a big margin. On the head dataset, benefiting from multi-domain learning, UOD achieves an MRE of $2.43mm$ and an SDR of 86.49\% within $4mm$, which is comparative with supervised method YOLO trained with at least 10 annotated labels, and much better than CC2D. On the hand dataset, there are some performance improvements in all metrics compared to CC2D, outperforming the supervised method YOLO trained with 25 annotated images. On the chest dataset, UOD shows the superiority of DATR which eliminates domain preference and balances the performance of all domains. In contrast, the performance of YOLO on chest dataset suffers a tremendous drop when the available labels are reduced to 25, 10, and 5. Figure~\ref{fig_visual} visualizes the predicted landmarks by UOD and CC2D.

\begin{table}[t]
\centering
\caption{Ablation study of different components of our DATR. Base is the basic transformer block; $\text{MSA}_{Q^d}$ denotes the domain-adaptive self-attention and $D^d$ denotes the domain-adaptive diagonal matrix. In each column, the best results are in \textbf{bold}.}
\label{tab_ablation}
\resizebox{1\linewidth}{!}{
\begin{tabular}{lrrrrrrrrrrrrr}
\hline
\multirow{3}*{Transformer}  &\multicolumn{5}{c}{Head~\cite{ref_head}}  &\multicolumn{4}{c}{Hand~\cite{gertych2007bone}}  &\multicolumn{4}{c}{Chest~\cite{shiraishi2000development}}  \\
 &MRE$\downarrow$ &\multicolumn{4}{c}{SDR$\uparrow$ (\%)}  &MRE$\downarrow$ &\multicolumn{3}{c}{SDR$\uparrow$ (\%)}  &MRE$\downarrow$ &\multicolumn{3}{c}{SDR$\uparrow$ (\%)}\\
 &(mm)& 2mm   & 2.5mm & 3mm   & 4mm   &(mm)     & 2mm & 4mm & 10mm  &(mm)  & 2mm & 4mm & 10mm\\
\hline
(a) Base       &24.95 &2.02 &3.17 &4.51 &5.85  &9.83 &5.33 &16.79&58.64  &58.11 &0.37 &1.96 &3.85 \\
\hline
(b)  +$D^d$          &22.75 &2.13 &3.24 &4.61 &6.96  &7.52 &6.13 &20.66&68.43  &52.98 &0.59 &2.17 &4.68 \\
\hline
(c) +$\text{MSA}_{Q^d}$         &2.51 &49.29 &60.89 &72.17 &84.36  &2.72 &48.56 &80.44&94.38  &9.09 &12.00 &19.33 &74.00 \\

\hline
(d) +$\text{MSA}_{Q^d}$+$D^d$  &\textbf{2.43} &\textbf{51.14}&\textbf{62.37}&\textbf{74.40}&\textbf{86.49}  &\textbf{2.52} &\textbf{53.37}&\textbf{84.27}&\textbf{97.59} &\textbf{8.49} &\textbf{14.00}&\textbf{39.33}&\textbf{76.33}\\
\hline
\end{tabular}
}
\end{table}

\noindent{\textbf{Ablation study:}}
We compare various components of the proposed domain-adaptive transformer. The experiments are carried out in UOD Stage II. As presented in Table~\ref{tab_ablation}, the domain-adaptive transformer has two key components: domain-adaptive self-attention $\text{MSA}_{Q^d}$ and domain-adaptive diagonal matrix $D^d$. The performances of (b) and (c) are much superior to those of (a) which demonstrates the effectiveness of $D^d$ and $\text{MSA}_{Q^d}$. Further, (d) combines the two components and achieves much better performances, which illustrates that domain-adaptive transformer improves the accuracy of detecting via cross-domain knowledge and global context information. We take (d) as the final transformer block.

\section{Conclusion}
To improve the robustness and reduce domain preference of multi-domain one-shot learning, we design a universal framework in that we first train a universal model via contrastive learning to generate pseudo landmarks and further use these labels to learn a universal transformer for accurate and robust detection of landmarks. UOD is the first universal framework of one-shot landmark detection on multi-domain data, which outperforms other one-shot methods on three public datasets from different anatomical regions. We believe UOD will significantly reduce the labeling burden and pave the path of developing more universal framework for multi-domain one-shot learning.
    
\bibliographystyle{splncs04}
\bibliography{paper297}
\end{document}